\documentclass[journal]{IEEEtran}
\ifCLASSINFOpdf
\else
\fi
\usepackage{ulem}
\usepackage{float}

\usepackage{algorithm}
\usepackage{algpseudocode}
\usepackage[marginal]{footmisc}
\usepackage[numbers,square,comma,sort&compress]{natbib}

\usepackage{graphics} 
\usepackage{epsfig} 
\usepackage{amsmath,amsfonts}
\usepackage{times} 
\usepackage{indentfirst}
\usepackage{bbding}
\usepackage{amssymb}  
\usepackage{booktabs}
\usepackage{url}
\usepackage{multirow}
\usepackage{hyperref}
\usepackage{xcolor}
\usepackage{blindtext}
\usepackage{mathtools}

\usepackage{etoolbox}
\makeatletter
\patchcmd{\@makecaption}
  {\scshape}
  {}
  {}
  {}
\makeatletter
\patchcmd{\@makecaption}
  {\\}
  {.\ }
  {}
  {}
\makeatother

\begin{document}
%
\title{Wavelet-based Multi-View Fusion of 4D Radar Tensor and Camera for Robust 3D Object Detection}
%
%
%

\author{Runwei Guan,
        Jianan Liu,
        Shaofeng Liang,
        Fangqiang Ding,
        Shanliang Yao,
        Xiaokai Bai,
        Daizong Liu, \\
        Tao Huang,~\IEEEmembership{~Senior Member,~IEEE},
        Guoqiang Mao,~\IEEEmembership{~Fellow,~IEEE},
        and Hui Xiong,~\IEEEmembership{~Fellow,~IEEE}
\thanks{Runwei Guan, Shaofeng Liang and Hui Xiong are with Thrust of Artificial Intelligence, Hong Kong University of Science and Technology (Guangzhou), Guangzhou, China. (Email: \{runwayrwguan, xionghui\}@hkust-gz.edu.cn; shawnliang123@163.com.)}
\thanks{Jianan Liu is with Momoniai AI, Goteborg, Sweden. (jianan.liu@momoniai.org)}
\thanks{Fangqiang Ding is with Department of Mechanical Engineering, Massachusetts Institute of Technology, Boston, USA. (fding@mit.edu)}
\thanks{Shanliang Yao is with School of Information Engineering, Yancheng Institute of Technology, Yancheng, China. (shanliang.yao@ycit.edu.cn)}
\thanks{Xiaokai Bai is with College of Information Science and Electronic Engineering, Zhejiang University, Hangzhou, China. (shawnnnkb@gmail.com)}
\thanks{Daizong Liu is with the Institute for Math \& AI, Wuhan University, Wuhan, China. (daizongliu@whu.edu.cn)}
\thanks{Tao Huang is with the College of Science and Engineering and the Centre for AI and Data Science Innovation, James Cook University, Smithfield, QLD 4878, Australia. (tao.huang1@jcu.edu.au)}
\thanks{Guoqiang Mao is with School of Transportation, Southeast University, Nanjing, China. (g.mao@ieee.org)}
\thanks{Corresponding author: xionghui@hkust-gz.edu.cn}
}

%
%

\markboth{IEEE Transactions on Multimedia, Submitted at Jan.~2026}%
{}
%



\maketitle

\begin{abstract}
4D millimeter-wave (mmWave) radar has been widely adopted in autonomous driving and robot perception due to its low cost and all-weather robustness. However, point-cloud-based radar representations suffer from information loss due to multi-stage signal processing, while directly utilizing raw 4D radar tensors incurs prohibitive computational costs. To address these challenges, we propose WRCFormer, a novel 3D object detection framework that efficiently fuses raw 4D radar cubes with camera images via decoupled multi-view radar representations. Our approach introduces two key components: (1) A Wavelet Attention Module embedded in a wavelet-based Feature Pyramid Network (FPN), which enhances the representation of sparse radar signals and image data by capturing joint spatial-frequency features, thereby mitigating information loss while maintaining computational efficiency. (2) A Geometry-guided Progressive Fusion mechanism, a two-stage query-based fusion strategy that progressively aligns multi-view radar and visual features through geometric priors, enabling modality-agnostic and efficient integration without overwhelming computational overhead. Extensive experiments on the K-Radar benchmark show that WRCFormer achieves state-of-the-art performance, surpassing the best existing model by approximately 2.4\% in all scenarios and 1.6\% in sleet conditions, demonstrating strong robustness in adverse weather.
\end{abstract}


%
\IEEEpeerreviewmaketitle

\section{Introduction}
\label{sec:intro}
%
%
%
%
\IEEEPARstart{R}obust perception is essential for autonomous driving and robotic navigation. Modern perception systems commonly rely on sensor suites, including cameras, LiDAR, and radar \cite{roy2022multi}. Cameras capture rich information such as color and texture, while LiDAR provides detailed three-dimensional (3D) geometric representations of the environment \cite{zhao2023lif,zhang2025synet,liu2023multi}. However, neither cameras nor LiDAR can directly measure object velocity, and both sensors are highly sensitive to illumination variations, particularly under adverse weather conditions such as rain, fog, or snow. In contrast, recent advances in 4D radar has shown enhanced abilities to capture vertical information compared to conventional 2D radars \cite{yao2023radar}. Despite this progress, many existing 4D radar perception methods based on point cloud representations remain constrained by sparsity, i.e., radar point clouds are notably sparse in 3D space compared to dense LiDAR scans \cite{liu2023smurf,jia2025radarnext,zhang2025dual,bi2025maff,SCKD}. Fortunately, radar and camera exhibit strong complementarity in object representation \cite{zheng2023rcfusion,xiong2023lxl,LXLv2,HGSFusion,bai2024sgdet3d,Doracamom}, making them a promising low-cost and efficient solution for robust perception.


\begin{figure}
    \centering
    \includegraphics[width=0.998\linewidth]{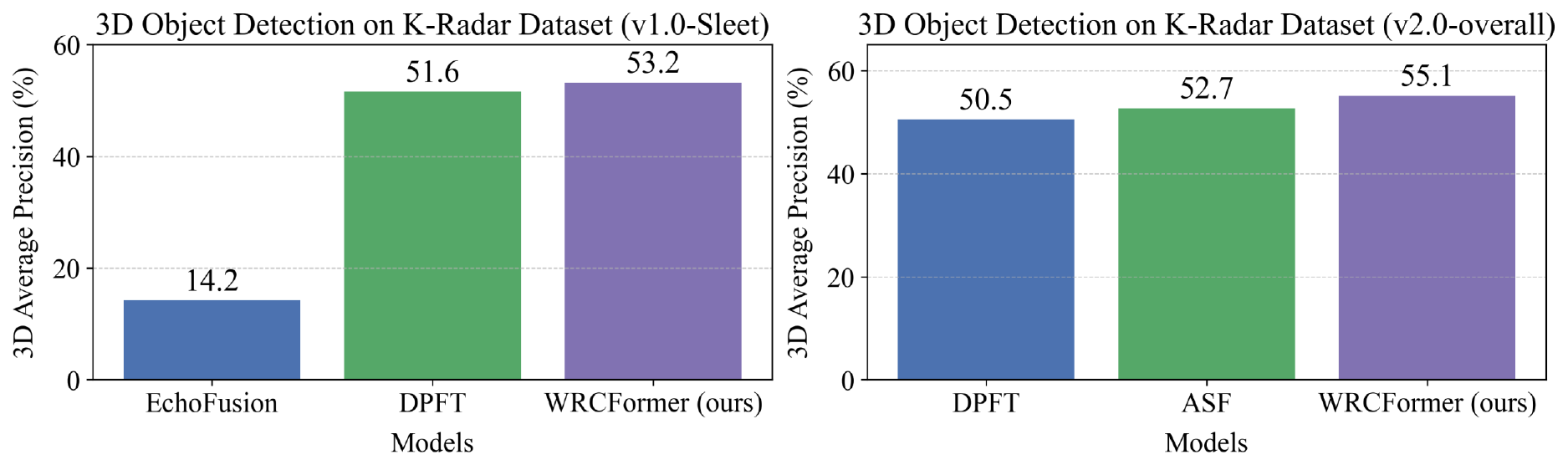}
    \vspace{-8mm}
    \caption{Overall comparison of state-of-the-art models, EchoFusion \cite{liu2023echoes}, DPFT \cite{fent2024dpft}, and ASF \cite{paek2025availability}, with WRCFormer (Ours) by fusion of camera and 4D radar tensors on K-Radar dataset (v1.0 and v2.0) \cite{paek2022k}.}
    \label{fig:overall_compare}
\end{figure}

In many radar-camera fusion models, radar data are commonly represented as point clouds. Technically, radar point clouds originate from the raw Analog-to-Digital Converter (ADC) signals received by the antenna array \cite{yao2025exploring}. Typically, a two-dimensional Fast Fourier Transform (2D FFT) is first applied to extract range and Doppler (velocity) information in the frequency domain. During this stage, side-lobe suppression techniques are often employed to enhance spectral clarity. Subsequently, Constant False Alarm Rate (CFAR) detection \cite{richards2005fundamentals} is applied to identify occupied range-Doppler bins while suppressing noise and false positives \cite{smith2021fcnn}. Finally, the Angle of Arrival (AoA) is estimated for each detected point, using methods such as FFT-based beam scanning, adaptive beamforming, subspace-based algorithms, or maximum likelihood techniques. As a result, the generated point clouds contain 3D spatial information, velocity, and Radar Cross Section (RCS). 

\begin{figure}
    \centering
    \includegraphics[width=0.87\linewidth]{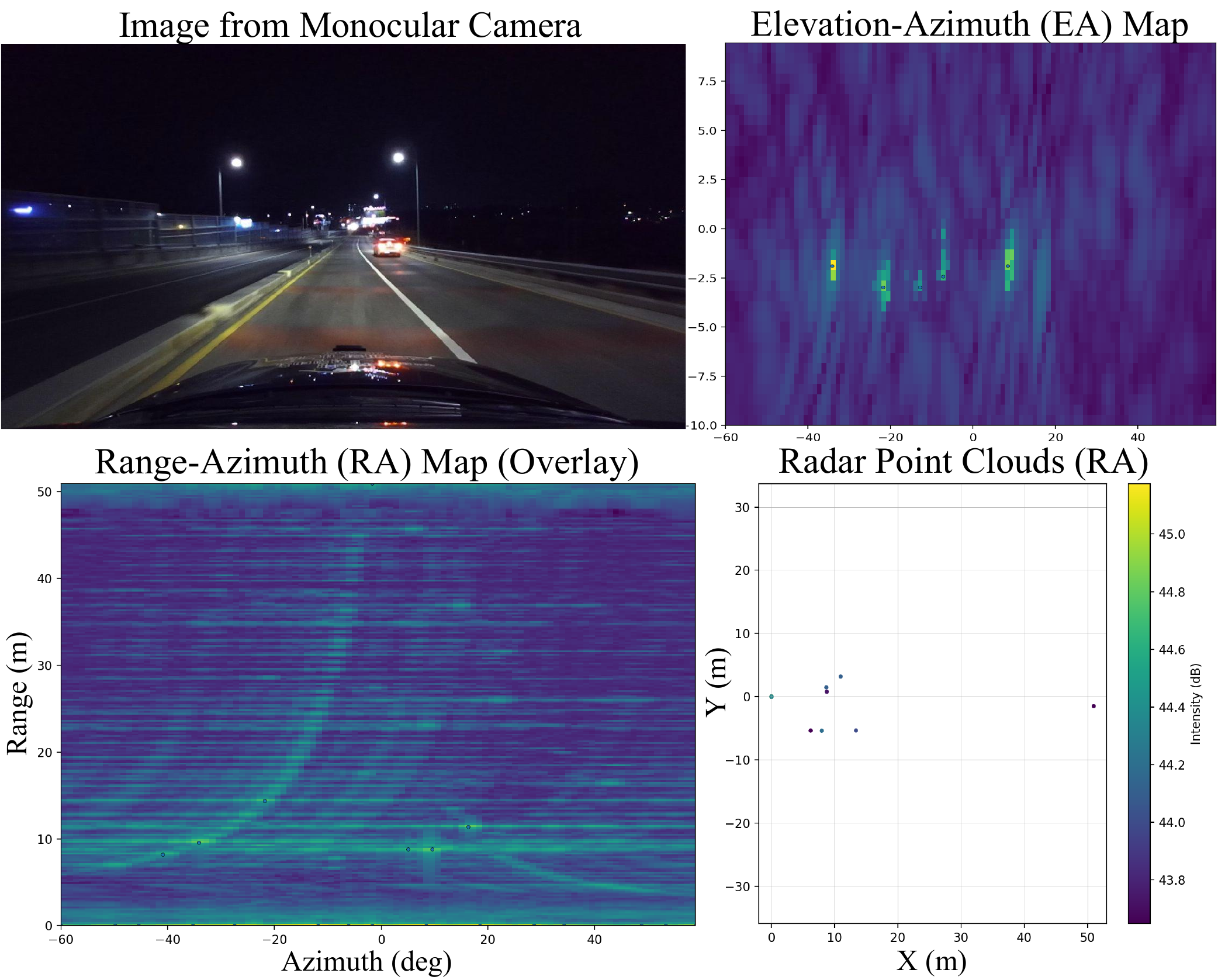}
    \vspace{-3mm}
    \caption{Visualization of one-frame data, including an image, elevation-azimuth map, range-azimuth map and the radar point clouds (Cartesian in BEV).}
    \label{fig:radar_pc_vis}
\end{figure}

While point cloud representations are more compact and computationally efficient compared to raw signal representations, they are inherently sparse, as illustrated in Fig. \ref{fig:radar_pc_vis} shows. More importantly, a substantial amount of environmental context is discarded during the CFAR process \cite{richards2005fundamentals}, which severely limits detection accuracy and prevents the radar modality from fully contributing when fused with other sensors \cite{liu2023echoes}.


Recent studies on camera and 4D radar raw-signal fusion have shown promising 3D perception performance in adverse conditions \cite{liu2023echoes,fent2024dpft,yang2023adcnet,paek2025availability}. However, the high angular resolution of 4D radar imposes substantial computational burden due to its large raw tensor size. To preserve angular resolution while maintaining computational efficiency, we propose a novel vision-radar fusion framework for 3D object detection. Following \cite{fent2024dpft}, we first decompose the 4D radar tensor into two complementary representations: a range-azimuth map and an elevation-azimuth map, each retaining full 4D attributes, including velocity, azimuth, elevation, range. 

Conventional convolution, while effective in many vision tasks, often struggle to capture the multi-scale and oscillatory patterns inherent in raw radar tensors. This limitation stems from their fixed local receptive fields and isotropy, which can blur subtle target boundaries and inadequately suppress spatially distributed clutter noise in radar data \cite{zhang2023peakconv}. In contrast, wavelet transforms provide a natural and mathematically grounded framework for analyzing signals with joint localization in space and frequency. This property is particularly aligned with the physical nature of radar sensing: scattering events manifest as localized energy concentrations across specific frequency bands (e.g., Doppler bins), while noise and clutter often spread across scales. By deploying learnable wavelet-based modules, our method explicitly decomposes the radar signal into multi-resolution sub-bands. This allows the network to strengthen foreground signatures by concentrating on relevant frequency components while effectively suppressing background clutter by attenuating high-frequency noise and diffuse interference, directly addressing the limitations of conventional convolutions for radar tensor representation.

In summary, our contributions are as follows:

\begin{itemize}
\item \textbf{WRCFormer}, an end-to-end 3D detector that establishes a new paradigm for fusing 4D radar tensors with multi-view imagery. By jointly encoding radar range-Doppler-azimuth-elevation cues and visual features in a unified bird’s-eye-view space, WRCFormer sets a new state-of-the-art on the K-Radar benchmark, outperforming existing methods by \textbf{3.1 to 8.7\ mAP} in adverse conditions. 
\item \textbf{Wavelet Attention Mixture-of-Experts (WA-MoE) Module}, a lightweight yet expressive radar feature extractor that leverages learnable wavelet decomposition coupled with sparse MoE gating. WA-MoE adaptively re-weights spatial–frequency components within the radar tensor, yielding discriminative representations that seamlessly integrate into FPN stages.

\item \textbf{Geometry-guided Progressive Fusion (GPF)}, a query-centric, modality-agnostic attention mechanism that iteratively aligns multi-view radar and image tokens via geometry-aware positional priors. 
\end{itemize}

\section{Related Works}
\label{sec:related}
\subsection{3D Object Detection with Radar Point Cloud}
Millimeter-wave (mmWave) radar serves as an all-weather perception sensor capable of measuring the 3D position, velocity, and RCS of objects. Recent advances in 4D mmWave radar, which provides denser point clouds than conventional 2D radar, have enabled standalone 3D object detection through deep learning models based on 3D voxel representations \cite{liu2023smurf, jia2025radarnext,musiat2024radarpillars,yang2025radar,weng20254draddet,bi2025maff}. Meanwhile, radar-camera fusion has shown promising performance by exploiting the complementary strengths of both modalities in a cost-effective manner \cite{yao2023radar,yao2025exploring}. Early fusion approaches typically combined radar point clouds with camera images for downstream perception tasks such as object detection \cite{xiong2023lxl,bai2024sgdet3d,zheng2023rcfusion,bi2025maff,zhao2024unibevfusion}. Another line of work explores radar-LiDAR fusion, where high-resolution geometric features from LiDAR complements radar features for improved detection performance \cite{wang2022interfusion,yang2024ralibev,song2024lirafusion,meng2024traffic}. Despite their effectiveness, point cloud representations remain inherently sparse and suffer from significant information loss due to multi-stage radar signal processing. To overcome this limitation, we propose an efficient fusion strategy that decomposes the raw 4D radar tensor into multi-view representations for integration with visual features, which preserves richer radar information while enabling robust radar-camera fusion for 3D object detection.

\subsection{3D Object Detection with Pre-CFAR Radar Data}
Previous studies, including approaches based on radar point clouds and tensor representations, have shown that radar can enhance 2D object detection \cite{nabati2021centerfusion} and depth estimation \cite{sun2025lircdepth}. However, conventional 2D radar lacks elevation resolution, limiting its ability to capture rich semantic and geometric cues, which constrains perception accuracy \cite{yao2025exploring}. This limitation has motivated a shift toward utilizing 4D radar tensors, which preserves the complete signal structure without projection-induced information loss. Research in this direction often employs pre-CFAR or raw radar data representations. Early efforts focus on fusing radar and camera data at the feature level for vehicle detection \cite{RA_R_C_early_paper}, while subsequent works directly processed 3D radar tensors (range-azimuth-Doppler) using deep learning for improved detection \cite{RA_R_early_paper}. To better exploit radar characteristics, researchers have proposed specialized network architectures, including RAMP-CNN for enhanced object recognition \cite{gao2020ramp}, RODNet for cross-supervised radar detection fused with camera localization \cite{RODNet}, and RaDDet for detecting dynamic road users \cite{zhang2021raddet}. Further advancements include Radatron, which uses cascaded multi-resolution radar inputs \cite{madani2022radatron}, and transformer-based models like T-FFTRadNet operating on raw ADC signals \cite{giroux2023t}. Researchers have also addressed efficient online detection using recurrent CNNs \cite{Recurrent_CNN_Raw_Radar} and sparse networks like SparseRadNet \cite{SparseRadNet}. More recently, methods including RTNH+ \cite{kong2024rtnh+} and SpikingRTNH \cite{paek2025spikingrtnh} have further advanced 4D radar object detection through improved pre-processing strategies and spiking neural network architectures. Several datasets support this line of research. RADIal \cite{rebut2022raw} provides ADC signals, range-azimuth maps, and point clouds with segmentation annotations. Other datasets, including VoD \cite{palffy2022multi}, TJ4DRadSet \cite{zheng2022tj4dradset}, Dual Radar \cite{zhang2025dual}, MAN TruckScenes \cite{fent2024mantruck}, and OmniHD-Scenes \cite{OmniHDScenes}, primarily offer radar point clouds. In contrast, our work builds upon the K-Radar dataset \cite{paek2022k}, which is currently the only public 4D radar \textit{tensor} dataset featuring multiple object types across diverse environments alongside corresponding 3D bounding box annotations. This unique combination enables a comprehensive evaluation of tensor-based 3D detection methods.

\begin{figure*}
    \centering
    \includegraphics[width=0.99\linewidth]{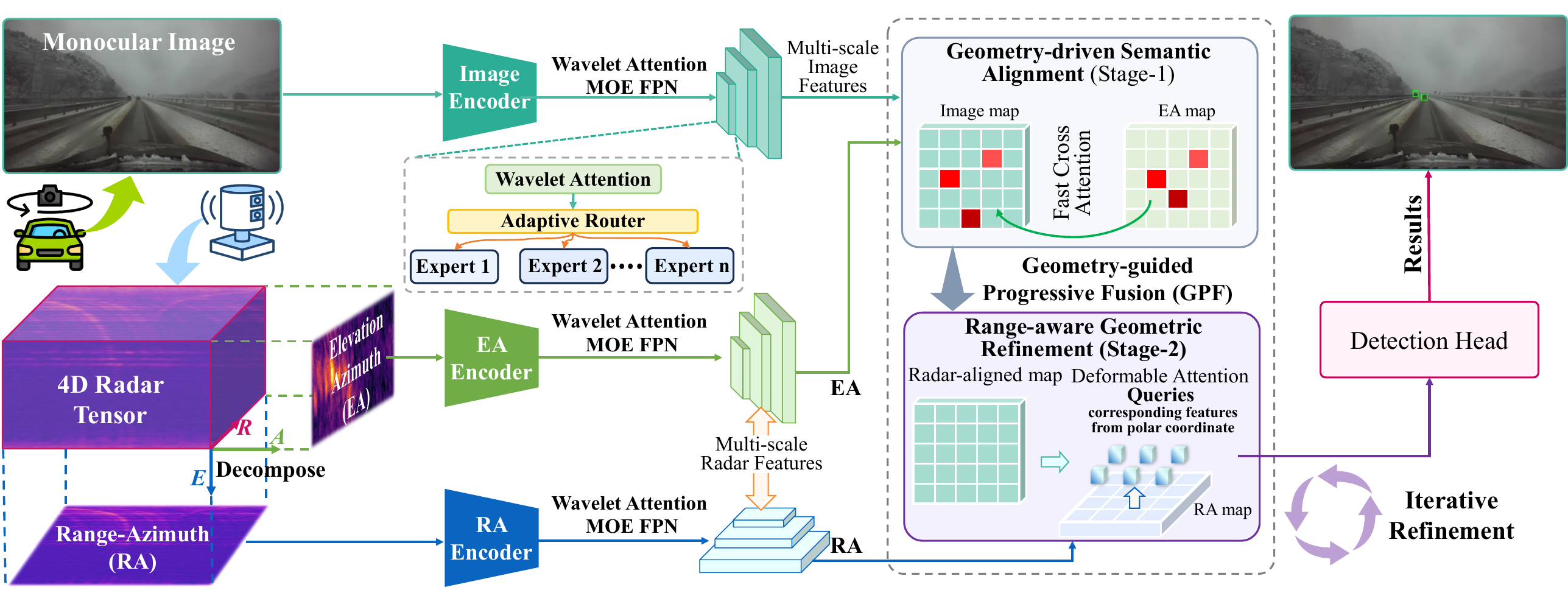}
    \vspace{-3mm}
    \caption{The architecture of proposed WRCFormer, which has two inputs: one-frame monocular image and one-frame 4D radar tensor. WRCFormer has one output of 3D bounding box.}
    \label{fig:model}
\end{figure*}

\section{Method}
We propose WRCFormer, a novel architecture designed for robust sensor fusion that processes inputs from a monocular camera and a 4D radar tensor as shown in Fig. \ref{fig:model}. The framework begins with a parallel feature extraction pipeline, in which an Image Encoder and a Feature Pyramid Network (FPN) generate multi-scale visual features. Meanwhile, to reduce the resolution mismatch between image and radar data, we adopt a 4D radar tensor (cube) and decompose it into Elevation-Azimuth (EA) and Range-Azimuth (RA) representations. These radar maps are fed into specialized encoders and an FPN based on our proposed Wavelet Attention Mixture-of-Experts (WA-MoE) module, which decomposes them into complementary high- and low-frequency components and adaptively fuses them. To efficiently integrate multi-view features from both modalities while addressing the semantic–geometric misalignment often overlooked in existing fusion strategies, we introduce a Geometry-guided Progressive Fusion (GPF) module as the core fusion component of WRCFormer. GPF performs a two-stage coarse-to-fine fusion: In the first stage, Geometry-driven Semantic Alignment (GSA) aligns semantic image features with geometric EA radar maps via fast cross-attention. In the second stage, Range-aware Geometric Refinement (RGR) further refines the aligned visual features using deformable attention queries guided by the RA radar map. Finally, the fused features are forwarded to a detection head that applies iterative refinement to progressively improve predictions and produce the final 3D detection results.

\subsection{Input Data of Image and 4D mmWave Radar}
From the perspective of perception sensors, a monocular camera captures RGB information of the environment projected onto a 2D image plane, whereas a 4D radar acquires environmental features in a polar coordinate system, namely the range–azimuth plane, which naturally aligns with the Bird’s-Eye View (BEV) and remains orthogonal to the image plane. At first glance, these two representations appear to share little common structure. However, the 4D radar tensor also provides elevation-azimuth perspectives that are consistent with the image plane, which facilitates cross-modal fusion. Nevertheless, directly operating on the 4D radar tensor incurs substantial computational costs due to its four-dimensional structure, including range, azimuth, elevation, and Doppler. Conversely, adopting sparse point cloud representations remains suboptimal, as excessive signal processing discards a significant amount of valuable information, as discussed in Section \ref{sec:intro}.

Motivated by these observations, we project the 4D radar tensor onto the range–azimuth (RA) and elevation–azimuth (EA) planes. Since projection inherently reduces dimensionality, we take measures to mitigate information loss and suppress noise. Following \cite{fent2024dpft}, we first define a set of 30 initial radar features, which prior studies have shown to be highly informative for radar-based perception tasks \cite{schumann2017comparison,scheiner2018radar}. We then train models using all 30 features and subsequently analyze the first-layer weights to assess the contribution of individual features to model convergence. In addition, we conduct sensitivity analysis by injecting noise into individual input features and monitoring the resulting output variations to evaluate the model’s robustness to input perturbations. Based on these analyses, we select the maximum, median, and variance of both amplitude and Doppler values as key features during radar data projection. Furthermore, the first and last three cells of the radar cube are truncated to suppress projection artifacts in the EA view.

\subsection{Feature Extraction}

The three distinct data representations, including RGB image, range-azimuth map, and elevation-azimuth map, are processed in parallel by three separate encoders to obtain high-dimensional, semantically rich features. To ensure fair and direct comparison with established baselines, we adopt the same encoder backbone configuration as in \cite{fent2024dpft}: a ResNet-101 for the RGB image stream and ResNet-50 backbones for the two radar tensor map streams.

Following the encoders, a custom-designed Feature Pyramid Network (FPN) is attached to each branch. Each FPN is augmented with a Wavelet Attention Mixture-of-Experts (WA-MoE) module, which fuses multi-scale features by selectively combining high-level contextual semantics with fine-grained spatial details, thereby enhancing detection robustness across scales. The specific design and operation of the WA-MoE module are elaborated in the following subsection.

\begin{figure}
    \centering
    \includegraphics[width=0.98\linewidth]{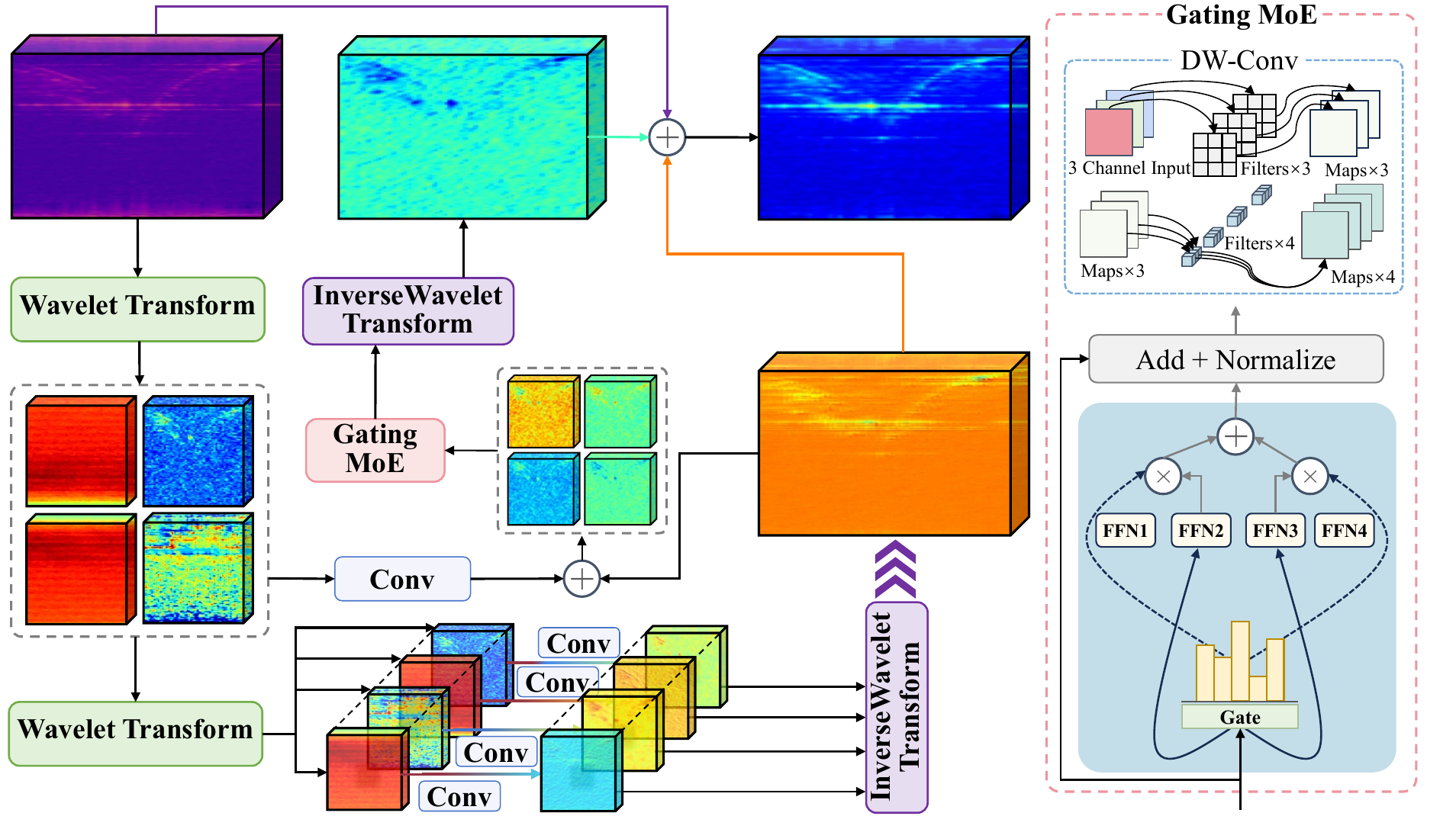}
    \vspace{-3mm}
    \caption{The structure of Wavelet Attention MoE (WA-MoE).}
    \label{fig:wa_moe}
\end{figure}

\subsection{Wavelet Attention Mixture-of-Experts}
To better capture this structure, we design a wavelet-based neural network for frequency-domain decomposition. Among various wavelet bases, we adopt the Haar wavelet due to its computational simplicity, orthogonality, and suitability for representing piecewise-constant signals commonly encountered in radar amplitude maps. The Haar transform also aligns with the binary and multi-resolution nature of digital signal processing, making it efficient for real-time deployment. Our design is further motivated by the following aspects that collectively address the limitations of conventional convolution-based FPNs in radar tensor processing:
\textbf{(1)} Noise components predominantly concentrate in the high-frequency sub-bands, whereas reflection peaks remain preserved in the low-frequency subbands. This property naturally facilitates target-clutter separation; 
\textbf{(2)} Radar sensing fundamentally operates in the frequency domain via Doppler and delay sampling, and wavelet tranform aligns well with the underlying radar scattering mechanism; 
\textbf{(3)} The introduction of a gated MoE is functionally equivalent to an adaptive filter bank, where the optimal spectral weights are estimated online for each channel and spatial location. This design enables data-driven spectrum shaping and offers greater flexibility than manually designed band-pass filters, while providing finer granularity than statically weighted Wavelet-CNNs; 
\textbf{(4)} For visual inputs, the high-frequency components of wavelets explicitly capture object edges, which further facilitate cross-modal alignment during the subsequent fusion process. 

Based on these insights, we propose the WA-MoE within the FPN as an efficient pre-processing stage prior to vision-radar feature fusion. Given an input radar feature map, denoted as $\mathbf{X} \in \mathbb{R}^{C\times H\times W}$, where $C$ represents the channel number and $H,W$ denote the spatial dimensions. $\mathbf{X}$ is first processed by a 2D Haar Discrete Wavelet Transform, $\mathtt{DWT}_{\mathtt{Haar}}(\cdot)$, as follows:
\begin{equation}
    \{\mathbf{X}_\mathrm{LL}, \mathbf{X}_\mathrm{LH}, \mathbf{X}_\mathrm{HL}, \mathbf{X}_\mathrm{HH}\} = \mathtt{DWT}_{\mathtt{Haar}}(\mathbf{X}),
\end{equation}
where $\mathbf{X}_\mathrm{LL}$ denotes the low-frequency approximation component, which captures the macroscopic scene structure and concentrates the main energy of the signal. $\mathbf{X}_\mathrm{LH}$ (horizontal details) captures the low-frequency information in the vertical direction and the high-frequency information along the horizontal direction, corresponding to the horizontal edges and textures in the signal. $\mathbf{X}_\mathrm{HL}$ (vertical details) captures the low-frequency information along the horizontal direction and the high-frequency information along the vertical direction, corresponding to the vertical edges and textures in the signal. $\mathbf{X}_\mathrm{HH}$ (diagonal details) denotes the high-frequency information along the diagonal direction, corresponding to the high-frequency noise and complex textures.

To extract informative representations from the wavelet domain at different levels of granularity, the architecture employs two parallel convolutional branches, which allows for the concurrent extraction of features from multiple orders of frequency decomposition. 

For the first branch corresponding to first-order wavelet domain encoding, the four sub-bands $\{\mathbf{X}_\mathrm{LL}, \mathbf{X}_\mathrm{LH}, \mathbf{X}_\mathrm{HL}, \mathbf{X}_\mathrm{HH}\}$ are fed into a convolutional module $\mathtt{Conv_1(\cdot)}$:
\begin{equation}
    F_1 = \mathtt{Conv_1}(\mathbf{X}_\mathrm{LL} \oplus \mathbf{X}_\mathrm{LH} \oplus \mathbf{X}_\mathrm{HL} \oplus \mathbf{X}_\mathrm{HH}),
\end{equation}
where $\oplus$ denotes concatenation operation and $F_1$ is the first-order wavelet domain encoding result. This branch directly extracts spatial correlations within each frequency sub-band from the first-level wavelet decomposition. Consequently, it captures explicit and prominent spatio-frequency patterns, serving as a shallow feature extractor in the wavelet domain.

For the second branch corresponding to second-order wavelet domain encoding, the sub-bands from the first step undergo another $DWT$, are processed by a second convolutional module $\mathtt{Conv_2}$, and finally reconstructed via an Inverse Wavelet Transform ($IWT$):
\begin{equation}
    F_2 = \mathtt{IWT_{Haar}}(\mathtt{Conv_2}(\mathtt{DWT_{Haar}}(\mathbf{X}_\mathrm{LL} \oplus \mathbf{X}_\mathrm{LH} \oplus \mathbf{X}_\mathrm{HL}  \oplus \mathbf{X}_\mathrm{HH}))).
\end{equation}

We integrate and refine the features extracted from the parallel branches using additive fusion to combine complementary feature sets. The resulting feature map $F_{\text{fused}}$ is then passed to a sparse Mixture-of-Experts (MoE) module, which consists of a gating network and $N$ expert networks. Instead of using a single, monolithic model to process all feature patterns, the MoE's gating network dynamically routes different feature patterns to specialized experts:
\begin{align}
     & F_{\text{fused}} = F_1 + F_2, \\
     & F_{\text{MoE}} = \sum_{i=1}^{N} \mathtt{g_i}(F_{\text{fused}}) \cdot \mathtt{E_i},
\end{align}
where $\mathtt{g_i}(F_{\text{fused}})$ represents the sparse gating function, which dynamically computes a weighted combination of experts $\mathtt{E_i(\cdot)}$ for an input. Each expert is implemented as a lightweight Feed-Forward Network (FFN) using the depthwise-separable convolutions and specializes in different patterns. 

The final stage transforms the processed features from the wavelet domain back to the signal domain and integrates them with the original information pathway. Specially, the output of the MoE module, $F_{\text{MoE}}$, is reconstructed back to the spatial domain via an $\mathtt{IWT}$. This reconstructed feature is then added to a residual connection from an earlier stage in the network:
\begin{align}
    & Y_{\text{residual}} = \mathtt{IWT_{\text{Haar}}}(F_{\text{MoE}}), \\
    & Y_{\text{out}} = Y_{\text{residual}} + X.
\end{align}

This residual design simplifies the learning task, alleviates the vanishing gradient problem in deep networks, and preserves both the fidelity of the original signal and the enhancements introduced by deep feature extraction in the final output.

\subsection{Geometry-guided Progressive Fusion}

\begin{figure}
    \centering
    \includegraphics[width=0.998\linewidth]{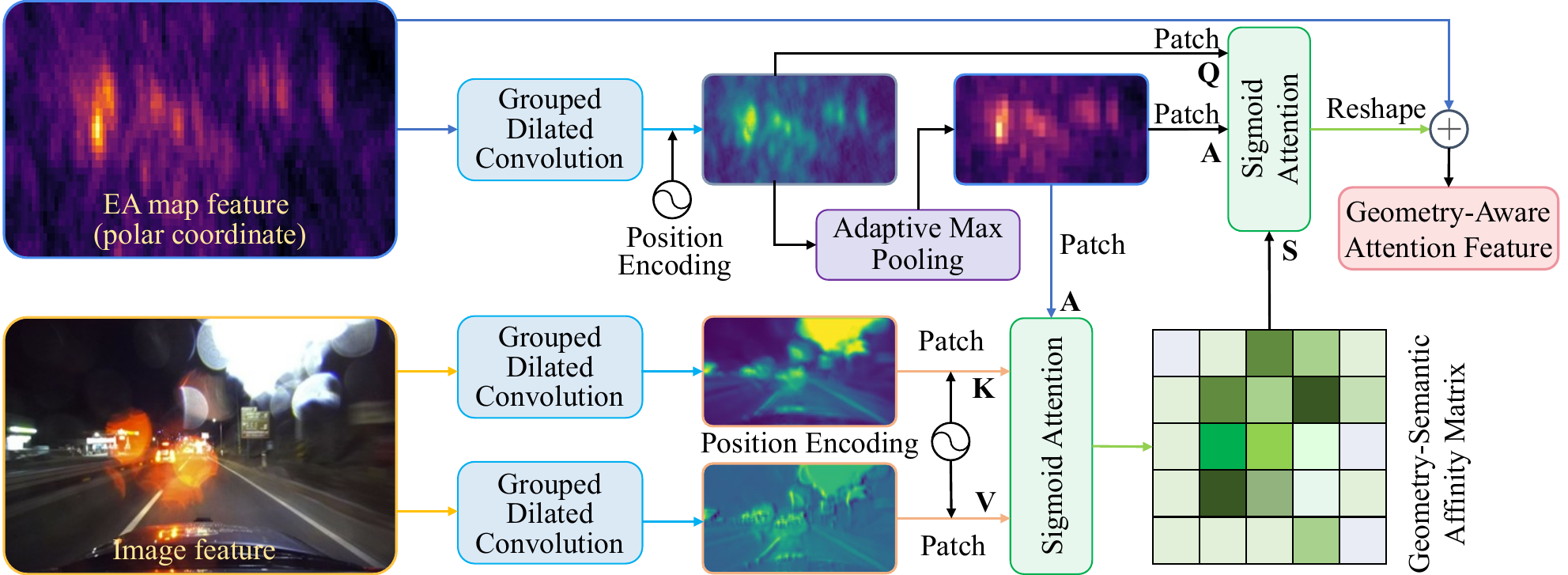}
    \vspace{-6mm}
    \caption{The structure of Geometry-driven Semantic Alignment, which is the stage-1 of Geometry-guided Progressive Fusion (GPF).}
    \label{fig:fast_cross_attn}
\end{figure}

To address the computational and alignment challenges in radar-camera fusion, we propose a Geometry-guided Progressive Fusion (GPF) module. GPF is a two-stage, coarse-to-fine fusion strategy designed to efficiently integrate multi-view radar and image features through structured geometric guidance, which includes Geometry-driven Semantic Alignment (GSA, Stage 1) and Range-aware Geometric Refinement (RGR, Stage 2).


In Stage 1, we apply grouped dilated convolutions follower by a dimension flatten operation to extract features from both the EA map $F^{EA} \in \mathbb{R}^{C\times H^{EA} \times W_{EA}}$ and the image representations $F^I \in \mathbb{R}^{C\times H_{I} \times W_{I}}$. Through this process, we obtain the query ($Q \in \mathbb{R}^{N \times C}$), key ($K \in \mathbb{R}^{M \times C}$) and value ($V \in \mathbb{R}^{M \times C}$), respectively. This design enlarges the receptive field while maintaining computational efficiency. Moreover, the use of dilated convolution facilitates the connection of sparse and discrete structures in the EA map, thereby enhancing the continuity of geometric priors within the feature space. The overall process is formulated as follows:
\begin{align}
\left\{
    \begin{aligned}
    & Q = \mathtt{Flat}(\mathtt{GDC}(F^{EA}) + PE), \\
    & K = \mathtt{Flat}(\mathtt{GDC}(F^I) + PE), \\
    & V = \mathtt{Flat}(\mathtt{GDC}(F^I) + PE),
    \label{eq:baca_1}
    \end{aligned}
    \right.
\end{align}
where $\mathtt{GDC}(\cdot)$ denotes the grouped dilated convolution and $PE$ indicates learnable position encoding. $\mathtt{Flat(\cdot)}$ indicates flattening along the spatial dimensions.

After that, we apply adaptive max-pooling to the EA query $Q \in \mathbb{R}^{N \times d}$ to obtain a compact representation $A \in \mathbb{R}^{n \times d}$, where $n \ll N$. Compared with adaptive average pooling, adaptive max pooling preserves the strongest signal responses in the EA features while effectively suppressing background interference, which preserves the most discriminative geometric responses while significantly reducing the sequence length.

Subsequently, we perform Sigmoid-based cross attention between $A$ and the image features $(K, V)$, yielding a geometry-semantic affinity matrix $S$:

\begin{equation}
    S = \sigma (\frac{QK^\mathrm{T}}{\sqrt{d_{qk}}}+b)V,
\end{equation}
where $\sigma(\cdot)$ is the Sigmoid function, $d_{qk}$ is the dimension of $Q$ and $K$, and $b$ is a learnable bias. This affinity matrix encodes coarse semantic–geometric correspondence under the guidance of the pooled EA representation.

Finally, the affinity matrix is used to refine the original EA queries by applying a second sigmoid cross attention between $(Q, A)$:

\begin{equation}
    F^{GS} = \sigma(\frac{QA^\mathrm{T}}{\sqrt{d_{qa}}}+b)S
\end{equation}
where $F^{GS}$ denotes the geometry-aware semantic feature. By replacing the direct quadratic interaction between $Q$ and $K$ with a two-step attention mechanism medicated by $A$, the overall complexity is reduced from $O(N\cdot |K|)$ to $O(n\cdot |K| + N \cdot n)$, which scales linearly with respect to the original length $N$.

\begin{figure}
    \centering
    \includegraphics[width=0.98\linewidth]{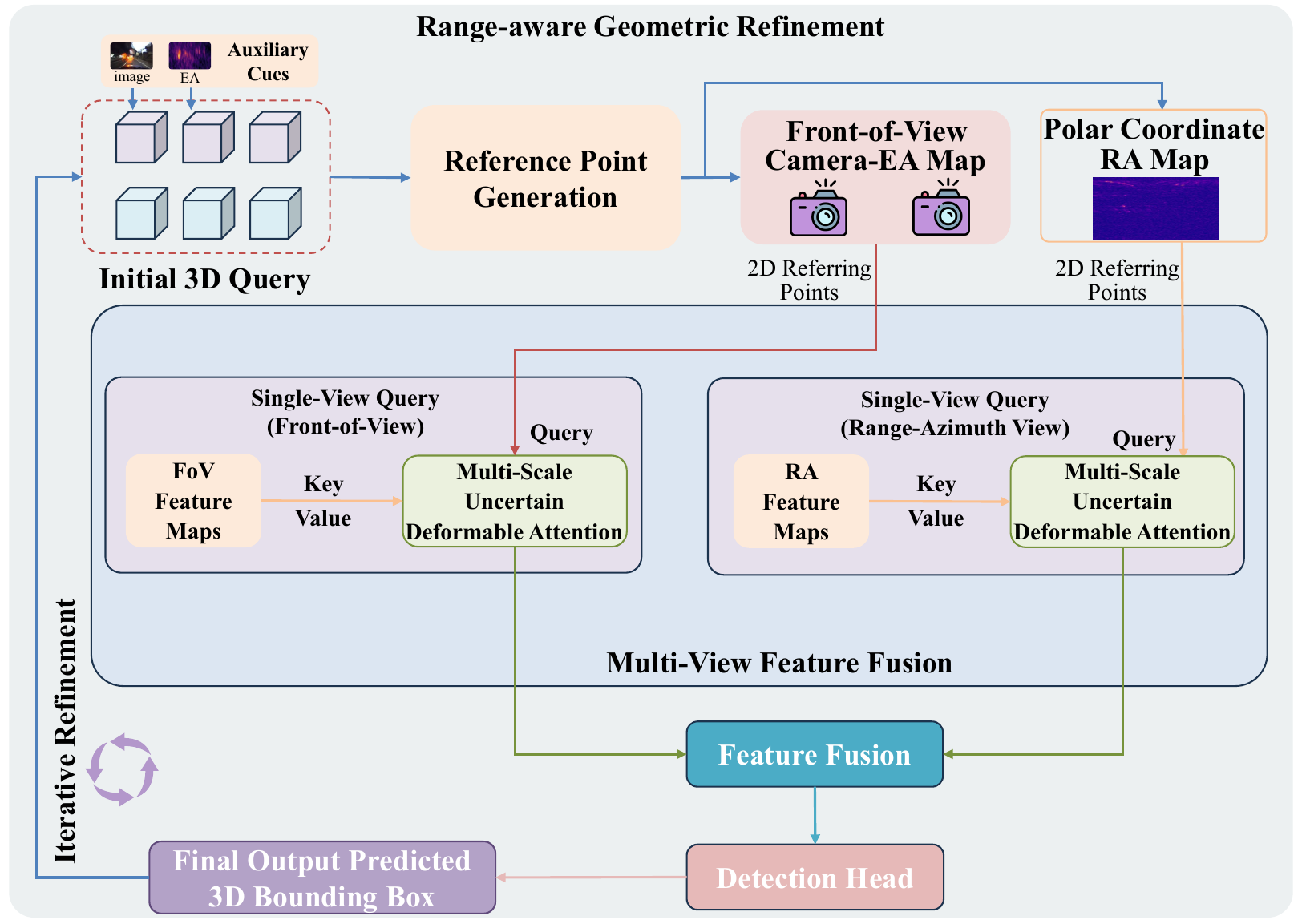}
    \vspace{-3mm}
    \caption{The structure of Range-aware Geometric Refinement, which is the stage-2 of Geometry-guided Progressive Fusion (GPF).}
    \label{fig:gpf-2}
\end{figure}

After Stage 1 produces the geometry-aware semantic features $F^{GS}$ (the Camera-EA fused output), Stage 2 further refines these queries by explicitly exploiting range priors from the RA map through a sequence of multi-modal reference point generation, dual-path uncertain deformable attention along the Camera–EA and RA path, feature fusion, and iterative detection refinement.

\textbf{Multi-modal guided reference-point generation}.
Rather than predicting reference points solely from the query embedding as in \cite{fent2024dpft}, we generate reference points using multi-modal guidance that fuses cues from the Stage-1 query, the image (camera) features, and the EA map. Concretely, for each inital query $Q^{(0)}_i$, we compute a fused embedding $\tilde Q_i$ as:
\begin{align}
    \tilde Q_i  = &\mathtt{LN}\Big( W_q Q^{(0)}_i + \mathtt{PoolAttn}\big(Q^{(0)}_i, F^I\big) \nonumber \\
    & + \mathtt{PoolAttn}\big(Q^{(0)}_i, F^{EA}\big)\Big),
\end{align}
where $\mathtt{PoolAttn}(\cdot,\cdot)$ denotes a lightweight, position-aware cross-attention or pooling operator that extracts spatial cues from image features ($F^I$) and EA features ($F^{EA}$) conditioned on $Q^{(0)}_i$, $W_q$ represents a learnable projection and $\mathtt{LN(\cdot)}$ denotes layer-normalization. From $\tilde{Q}_i$, we regress both the reference point offset and a confidence (uncertainty) score:
\begin{align}
\left\{
    \begin{aligned}
    & \Delta p_i = W_{\Delta}\tilde Q_i, \\
    & p_i = p^{\text{anchor}}_i + \Delta p_i, \\
    & c_i = \sigma(W_c \tilde Q_i),
    \label{eq:uncertain_1} \\
    \end{aligned}
    \right.
\end{align}
where $p^{anchor}_i$ denotes a coarse anchor (e.g. grid location or Stage-1 coarse projection), $p_i$ represents the predicted 2D reference coordinate in RA space, $c_i\in(0,1)$ is the learned confidence, and $\sigma(\cdot)$ denotes the sigmoid function.

\textbf{Camera-EA deformable uncertain attention (FoV path).}  
Using the reference points $p_i$, we apply a deformable, uncertainty-aware cross attention over front-of-view (FoV) multi-scale $\{F^{GS}_s\}^S_{s=1}$. For query $i$, the camera-path refined feature is
\begin{equation}
    \tilde F^{GS}_i=\sum_{s=1}^{S}\sum_{k=1}^{K}u^{\text{GS}}_{i,s,k}\ w^{\text{GS}}_{i,s,k} \ F^{GS}_s\big(p_i + \Delta p^{GS}_{i,s,k}\big),
\end{equation}
where $s$ feature pyramid levels, $K$ denotes number of sampling offsets per scale; $p^{GS}_{i,s,k}$ represents learnable sampling offsets relative to $p_i$; $w^{GS}_{i,s,k}$ denotes the normalized attention weights produced by a small projection of the query; $u^{GS}_{i,s,k} \in [0,1]$ denotes an uncertainty weight predicted per sample (or per offset) that down-weights unreliable samples caused by factors such as occlusion or sensor noise.

\textbf{RA multi-scale deformable uncertain attention (Range-Azimuth path).}
In parallel, the same reference points $p_i$ guide a multi-scale deformable attention over the RA feature pyramid $\{F^{RA}_s\}$. This path explicitly emphasizes explicit range constraints. For the $i$-th query, we compute the refined RA-path feature as
\begin{equation}
    \tilde F^{RA}_i=\sum_{s=1}^{S}\sum_{k=1}^{K}u^{RA}_{i,s,k}\ w^{RA}_{i,s,k}\ F^{RA}_s\big(p_i + \Delta p^{RA}_{i,s,k}\big),
\end{equation}
with analogous definitions for $p^{RA}_{i,s,k}$, $w^{RA}_{i,s,k}$ and uncertainty weights $u^{RA}_{i,s,k}$. Since RA measurements vary significantly with range, the RA path adopts multi-scale sampling to capture both near and far geometric cues.

\textbf{Feature Fusion.} The FoV-path and RA-path outputs are fused with the original query via concatenation and a learnable projection:
\begin{equation}
    \tilde Q_i = \mathtt{FFN}\big(\mathtt{Concat}(\tilde F^{GS}_i, \tilde F^{RA}_i\big),
\end{equation}
where $\mathtt{FFN}(\cdot)$ denotes a fully-connected layer. This design preserves the distinct contributions of each modality while allowing the network to learn an optimal joint embedding.

\begin{table*}
    \centering
    \caption{3D object detection prediction on K-Radar dataset (v2.0 version). \textbf{C} and \textbf{R} denote camera and radar, respectively. R$_\text{RA}$ and R$_\text{AE}$ are two view of radar tensor map (range-azimuth and azimuth-elevation).}
    \setlength\tabcolsep{3.8pt}
    \vspace{-3mm}
    \begin{tabular}{cc|ccccccc|ccccccc}
    \hline
       \multirow{2}[2]{*}{\textbf{Models}} &  \multirow{2}[2]{*}{\textbf{Venues (Year)}} & \textbf{C} & \textbf{R$_\text{AE}$} & \textbf{R$_\text{RA}$} & \textbf{R} & \textbf{C + R$_\text{EA}$} & \textbf{C + R$_\text{RA}$} & \textbf{C + R} & \textbf{C} & \textbf{R$_\text{EA}$} & \textbf{R$_\text{RA}$} & \textbf{R} & \textbf{C + R$_\text{AE}$} & \textbf{C + R$_\text{RA}$} & \textbf{C + R} \\
       \cmidrule(lr){3-9} \cmidrule(lr){10-16}
       & &  \multicolumn{7}{c}{\textbf{Sedan}} & \multicolumn{7}{c}{\textbf{Bus or Truck}} \\
    \hline
       DPFT & TIV$_{\text{2023}}$ & 8.9 & 4.4 & 35.0 & 36.2 & 11.1 & 48.5 & 50.5 & 5.2 & 2.6 & 24.8 & 28.2 & 7.3 & 29.8 & 33.6 \\
    \hline
      ASF & NIPS$_{\text{2025}}$ & \textbf{14.8} & - & - & - & - & - & 52.7 & 9.6 & - & - & - & - & - & 36.2\\
    \hline
      \textbf{WRCFormer} & \textbf{Ours$_{\text{2025}}$} & 12.1 & \textbf{8.4} & \textbf{42.0} & \textbf{45.9} & \textbf{16.7} & \textbf{53.8} & \textbf{56.4} & \textbf{9.8} & \textbf{6.3} & \textbf{31.7} & \textbf{35.6} & \textbf{10.1} & \textbf{33.5} & \textbf{38.7} \\
    \hline
    \end{tabular}  
    \label{tab:kradar_version_2}  
\end{table*}

\begin{table*}
\setlength\tabcolsep{9.2pt}
\caption{Overall comparison of average precision (AP) for 3D object detection on K-Radar (v1.0 version) with a threshold of 0.3.}
\vspace{-3mm}
\centering
\label{tab:vg_datasets}
\begin{tabular}{lc|cccccccccc}  
\hline  
    \multirow{2}[2]{*}{\textbf{Models}} & \multirow{2}[2]{*}{\textbf{Venues (Year)}} & \multirow{2}[2]{*}{\textbf{Sensors}} & \multirow{2}[2]{*}{\textbf{Normal}} & \multirow{2}[2]{*}{\textbf{Overcast}} & \multirow{2}[2]{*}{\textbf{Fog}} & \multirow{2}[2]{*}{\textbf{Rain}} & \multirow{2}[2]{*}{\textbf{Sleet}} & \textbf{Light} & \textbf{Heavy} & \textbf{Total} \\   
    & & &  &  & & & & \textbf{Snow} & \textbf{Snow} & \textbf{mAP$_{\text{3D}}$}  \\
    \hline
    RTNH & NIPS$_{\text{2022}}$ & R & 49.9 & 56.7 & 52.8 & 42.0 & 41.5 & 50.6 & 44.5 & 47.4 \\
    \hline
    Voxel R-CNN & AAAI$_{\text{2021}}$ & L & 81.8 & 69.6 & 48.8 & 47.1 & 46.9 & 54.8 & 37.2 & 46.4 \\
    CasA & TGRS$_{\text{2022}}$ & L & 82.2 & 65.6 & 44.4 & 53.7  & 44.9 & 62.7 & 36.9 &  50.9 \\
    TED-S & AAAI$_{\text{2022}}$ & L & 74.3 & 68.8 & 45.7 & 53.6 & 44.8 & 63.4 & 36.7 & 51.0 \\
    \hline
    VPFNet & TMM$_{\text{2023}}$ & C + L & 81.2 & 76.3 & 46.3 & 53.7 & 44.9 & 63.1 & 36.9 & 52.2 \\
    TED-M & AAAI$_{\text{2023}}$ & C + L & 77.2 & 69.7 & 47.4 & 54.3 & 45.2 & 64.3 & 36.8 & 52.3 \\
    MixedFusion & TNNLS$_{\text{2024}}$ & C + L & 84.5 & 76.6 & 53.3 & 55.3 & 49.6 & 68.7 & 44.9 & 55.1 \\
    \hline
    EchoFusion & NIPS$_{\text{2023}}$ & C + R & 51.5 & \textbf{65.4} & 55.0 & 43.2 & 14.2 & 53.4 & 40.2 & 47.4 \\
    DPFT & TIV$_{\text{2023}}$ & C + R & 55.7 & 59.4 & 63.1 & 49.0 & 51.6 & 50.5 & 50.5 & 56.1 \\
    \textbf{WRCFormer} & \textbf{Ours$_{\text{2025}}$} & C + R & \textbf{55.9} & 59.7 & \textbf{70.1} & \textbf{52.2} & \textbf{54.7} & \textbf{59.2} & \textbf{54.0} & \textbf{58.7} \\
\hline
\end{tabular}
\end{table*}


\textbf{Detection head and iterative refinement.} The fused representation $\hat{Q}_i$ is fed into the detection head \cite{fent2024dpft}, which performs object classification and 3D bounding box regression while updating the query for the subsequent refinement iteration. The query update follows a residual MLP pattern:
\begin{equation}
    Q^{t+1}_i = Q^t_i + \mathtt{FFN}(\mathtt{LN}(\hat{Q}_i))
\end{equation}
and the detection head produces predictions as
\begin{equation}
    (\hat{y}^t_i, \hat{b}^t_i) = \mathtt{Head}(Q^t_i),
\end{equation}
where $\hat{y}_i^t$ denotes class confidence and $\hat{b}^t_i$ represents the regressed 3D bounding box parameters. The process is iterated for $T$ refinement steps, progressively improving localization and confidence. The learned reference confidence $c_i$ and sampling uncertainties $u$ can be fused into the final detection score:
\begin{equation}
    s_i^t = \hat{y}_i^t \cdot c_i \cdot \frac{1}{K}\sum_{s,k} u_{i,s,k},
\end{equation}
which down-ranks detections relying on noisy or uncertain measurements.

\subsection{Training Objectives}
We follow the training objective of RT-DETR with the bipartite matching paradigm. Specifically, a Hungarian matching algorithm \cite{zhudeformable} is employed to establish one-to-one correspondences between the set of predicted objects and ground-truth annotations, where the matching cost is defined as a weighted combination of classification and bounding box regression errors.

Formally, the overall training loss is defined as:

\begin{equation}
    \mathcal{L} = \lambda_{\text{cls}}\mathcal{L}_{\text{cls}} + \lambda_{\text{box}} \mathcal{L}_{\text{box}}, 
\end{equation}
where $\mathcal{L}_{\text{cls}}$ denotes the classification loss and $\mathcal{L}_{\text{box}}$ the regression loss.

\section{Experiments}
\subsection{Implementation Details}
\textbf{Datasets.} We train and evaluate our method on the K-Radar dataset, which is currently the only large-scale 3D object detection dataset based on 4D radar tensors in conjunction with complementary sensor modalities such as cameras and LiDAR. K-Radar comprises approximately 35K multi-modal frames captured across diverse road environments, including urban, suburban, highway, and alleyway scenarios, under a wide range of weather conditions such as clear, foggy, rainy, sleet, and snowy scenes. Each frame contains 4D radar tensors, encompassing range, azimuth, elevation, Doppler dimensions, together along with carefully annotated 3D bounding boxes for 93.3K objects across five categories: sedan, bus/truck, pedestrian, bicycle, and motorcycle. In addition to radar measurements, the dataset provides high-resolution LiDAR point clouds, stereo RGB images, and RTK-GPS/IMU data for precise spatial and temporal calibration. The complete dataset occupies approximately 18 TB, reflecting its comprehensive multi-sensor coverage and high-fidelity data quality.

\textbf{Training.} We resize the size of image as 512 $\times$ 512 pixels. We adopt the cosine learning rate scheduler with the initial learning rate as 1 $\times$ 10$^{-4}$. We train the model for 200 epochs using a batch size of 4 on a single NVIDIA RTX 4090 GPU. We employ AdamW optimizer. For the encoders, we initialize the networks with weights pretrained on ImageNet.

\textbf{Models.} We include three categories of models: (1) radar tensor-based models (RTNH), (2) LiDAR point cloud-based models (Voxel R-CNN, TED-S, CasA), (3) LiDAR-camera fusion models (MixedFusion, VPFNet, TED-M), and (4) radar–camera fusion models (EchoFusion, DPFT, ASF).

\textbf{Evaluation.} We evaluate all models on revision v1.0 and revision v2.0, where revision v2.0 includes corrected object heights and previously missing object annotations. We adopt average precision (AP) as the metric, including AP for BEV and 3D bounding box prediction, with the Intersection of Union (IoU) threshold above 0.3. 

\subsection{Quantitative Results}

\begin{figure*}
    \centering
    \includegraphics[width=0.95\linewidth]{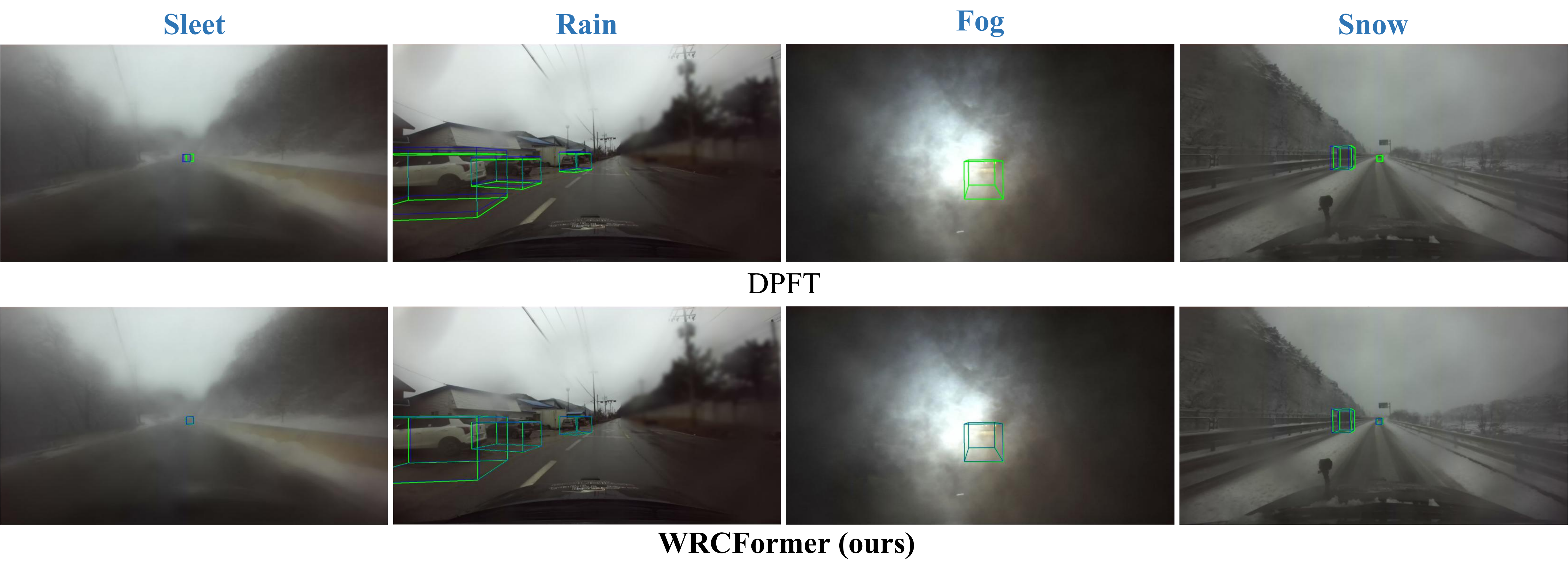}
    \vspace{-3mm}
    \caption{Visualization of predicted results by WRCFormer and DPFT on K-Radar dataset, where blue bounding box denotes predicted results while green bounding box denotes ground truth.}
    \label{fig:overall_vis}
\end{figure*}

\textbf{Comparison of performances upon different input modalities.} Building upon the experimental protocols established in prior studies \cite{fent2024dpft,paek2025availability},  we benchmark our proposed WRCFormer against DPFT and ASF under diverse sensor input configurations on the K-Radar dataset with v2.0 annotations. The results reported in Table \ref{tab:kradar_version_2} indicate that WRCFormer consistently outperforms both ASF and DPFT across all evaluated settings. Furthermore, we observe that the Range-Azimuth map derived from radar data contributes more significantly to object detection performance than the Elevation-Azimuth map. Additionally, radar-only detection demonstrates substantially stronger robustness under adverse weather conditions compared to camera-only approaches.

\textbf{Overall Performances (v1.0 version).} WRCFormer achieves the highest Total mAP3D of 58.7 (Table \ref{fig:overall_compare}), outperforming camera-LiDAR fusion methods. This gain stems from our WA-MoE module and GPF mechanism, which enhance sparse radar representations and enable robust cross-modal alignment. In clear conditions (Normal/Overcast), camera-LiDAR fusion excels, benefiting from dense geometry. Conversely, in adverse weather (Fog, Rain, Sleet, Snow), our camera-radar fusion significantly outperforms camera-LiDAR alternatives, validating the effectiveness of wavelet-based frequency decomposition and geometry-guided fusion under sensor degradation. WRCFormer also surpasses other radar-camera methods (DPFT, EchoFusion) across most conditions, with a minor shortfall only against EchoFusion in Overcast weather, likely due to architectural differences in clear-sky feature extraction. The consistent advantage in challenging scenarios demonstrates the strength of our joint spatial-frequency encoding and progressive geometric alignment.


\begin{table}
    \centering
    \caption{Performances of basic modules for FPN on K-Radar (v1.0).}
    \setlength\tabcolsep{2.4pt}
    \vspace{-3mm}
    \begin{tabular}{c|cccc}
    \hline
      \textbf{Methods} & \textbf{mAP$_{\text{3D}}$} & \textbf{mAP$_{\text{BEV}}$} & \textbf{Params (K)} & \textbf{FLOPs (G)} \\
    \hline       CSPDarkNet & 54.4 & 55.7 & 328.45 & 33.712\\
      Deformable Conv v4 & 55.9 & 57.0 & - & 60.398 \\
      GhostNet  & 53.8 & 54.7 &  83.85 & 11.069 \\
      RepVGG & 54.1 & 55.2 &  656.89 & 67.42 \\
      \hline
      \textbf{Wavelet Attention MoE} & \textbf{58.7} & \textbf{59.4} & \textbf{40.72} & \textbf{2.674}\\ 
    \hline
    \end{tabular}  
    \label{tab:fpn_module}  
\end{table}

\textbf{Comparison of multi-scale feature representation.} As detailed in Table \ref{tab:fpn_module}, we compare the performance of different FPN building blocks. Our proposed Wavelet Attention MoE (WA‑MoE) module, leveraging its dynamic routing mechanism combined with a wavelet attention design that effectively highlights critical radar signal patterns, significantly outperforms other architectural blocks during inference. Specifically, it achieves an mAP$_\text{3D}$ value 2.8 points higher than Deformable Convolution v4, while maintaining lower FLOPs than several alternative building blocks.

\begin{figure}
    \centering
    \includegraphics[width=0.92\linewidth]{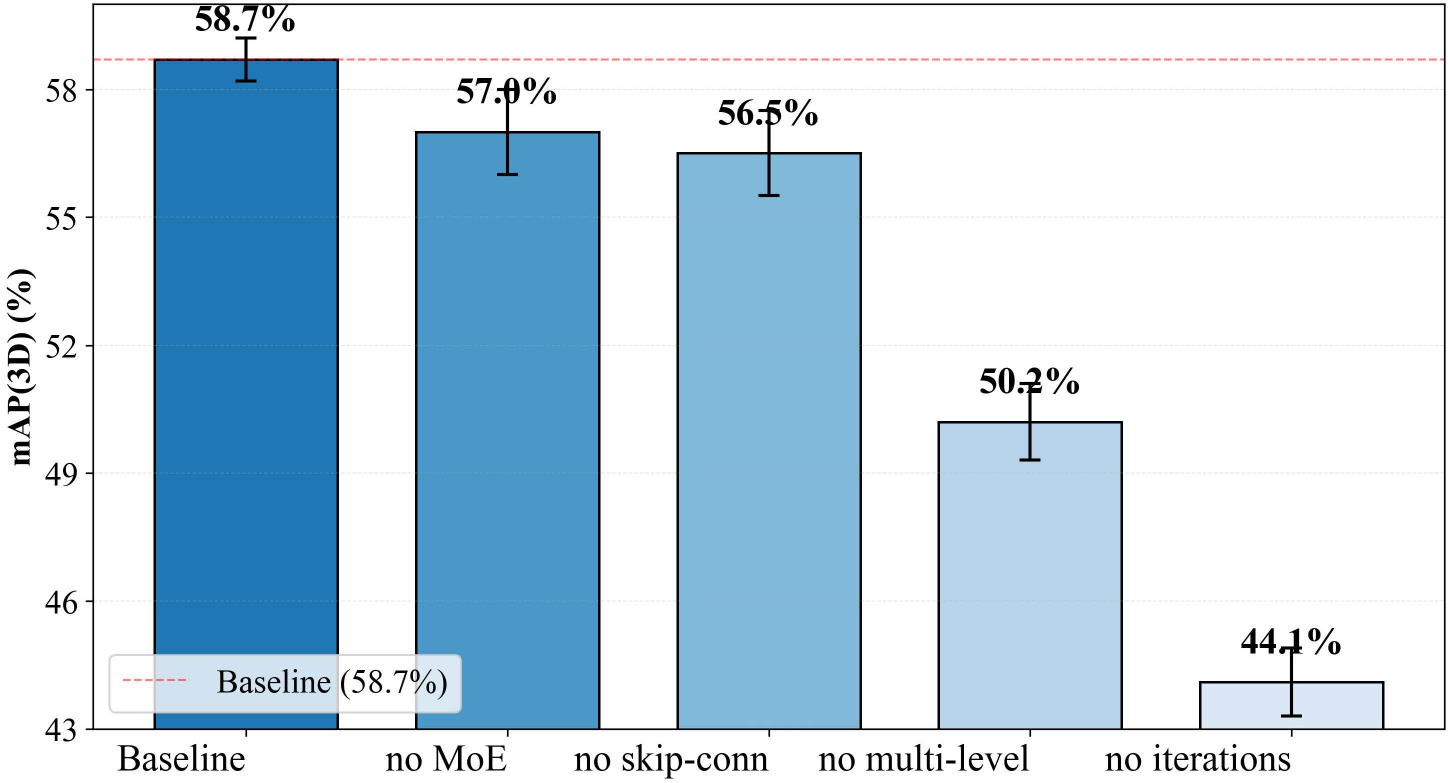}
    \vspace{-3mm}
    \caption{Ablation results on individual settings in WRCFormer.}
    \label{fig:ablation_fig}
\end{figure}

\begin{table}
    \centering
    \caption{The comparison and ablation experiments of Geometry-guided Progressive Fusion (GPF) on K-Radar (v1.0).}
    \setlength\tabcolsep{9.5pt}
    \vspace{-3mm}
    \begin{tabular}{c|ccc}
    \hline
    \textbf{Methods} & \textbf{Params (M)} & \textbf{mAP$_{\text{3D}}$} & mAP$_{\text{BEV}}$ \\
    \hline
      IMPFusion \cite{fent2024dpft}  & 0.293 & 56.6 & 57.2  \\
    \hline
      GPF (CA+RGR) & 0.795 & 58.1 & 58.8 \\
      GPF (no uncertainty) & \textbf{0.179} & 57.4 & 58.1 \\
    \hline
    \multicolumn{4}{c}{$A \in \mathbb{R}^{12 \times 12}$} \\
    \hline
      \textbf{GPF (GSA+RGR)} & 0.208 & \textbf{58.7} & \textbf{59.4} \\
    \hline
    \multicolumn{4}{c}{$A \in \mathbb{R}^{8 \times 8}$} \\
    \hline
      \textbf{GPF (GSA+RGR)} & 0.192 & 57.6 & 58.4 \\
    \hline
    \end{tabular}
    \label{tab:placeholder}
\end{table}

\textbf{Comparison of methods for camera-radar fusion.} As shown in Table IV, our proposed Geometry-guided Progressive Fusion (GPF) consistently outperforms IMPFusion in terms of detection accuracy while maintaining a reduced parameter count. Specifically, GPF achieves a higher mAP while using fewer parameters. When we replace the Geometry-driven Semantic Alignment (GSA) with a vanilla cross‑attention mechanism, the parameter size increases notably, yet the resulting mAP remains lower than that achieved by GPF. Furthermore, removing the uncertainty weighting from the Range‑aware Geometric Refinement (RGR) module leads to a clear performance drop, highlighting its importance for robust fusion. Finally, we examine the effect of varying the size of the adjacency matrix $A$ in GSA, which serves as a tunable hyper‑parameter. Reducing the size of $A$ results in a slight decrease in mAP, but also yields a more compact model. This indicates that GPF offers a flexible trade‑off between accuracy and model complexity by adjusting the dimension of $A$.

\begin{table}
    \centering
    \caption{Comparison of memory usage and FPS of models.}
    \setlength\tabcolsep{13.0pt}
    \vspace{-3mm}
    \begin{tabular}{c|c|c|c}
    \hline
      \textbf{Models}  & \textbf{Sensors} & \textbf{VRAM (GB)} & \textbf{FPS} \\
    \hline
       3D-LRF \cite{chae2024towards}  &  L + R & 1.2 & 5.04      \\
       ASF \cite{paek2025availability}     &  C + L + R & 1.6 & 13.5  \\
       DPFT \cite{fent2024dpft}    &  C + R & 2.6 & 10.8      \\
    \hline
       \textbf{WRCFormer} & \textbf{C + R} & \textbf{2.4} & \textbf{14.8} \\
    \hline
    \end{tabular}
    \label{tab:memory}
\end{table}

\textbf{Comparison of computational efficiency.} Table \ref{tab:memory} shows the memory usage and FPS of different models on a single RTX 3090 GPU. WRCFormer consumes approximately 2.4 GB of GPU memory, which is slightly lower than DPFT (2.6 GB) but remains higher than ASF (1.6 GB) and 3D-LRF (1.2 GB). In terms of inference speed, WRCFormer achieves the fastest throughput, reaching 14.8 frames per second (FPS).

\textbf{Ablation studies on other settings in WRCFormer.} As Fig. \ref{fig:ablation_fig} shows, we conduct comprehensive ablation studies on several key components of the proposed framework: the Mixture‑of‑Experts (MoE) module in WA‑MoE, the skip‑connection design in FPN, the multi‑level feature configuration of FPN, and the iterative refinement mechanism in the detection head. Experimental results confirm that each of these design choices plays an essential role and contributes significantly to the overall detection performance.

\subsection{Qualitative Results}
Fig. \ref{fig:overall_vis} presents the qualitative detection results on the K-Radar dataset, comparing our WRCFormer with DPFT. It can be observed that under various adverse weather conditions, including sleet, rain, fog, and snow, WRCFormer achieves fewer missed detections and exhibits stronger robustness against environmental disturbances compared to DPFT. Furthermore, WRCFormer demonstrates superior capability in detecting small-scale objects under these challenging scenarios.

\begin{figure}
    \centering
    \includegraphics[width=0.95\linewidth]{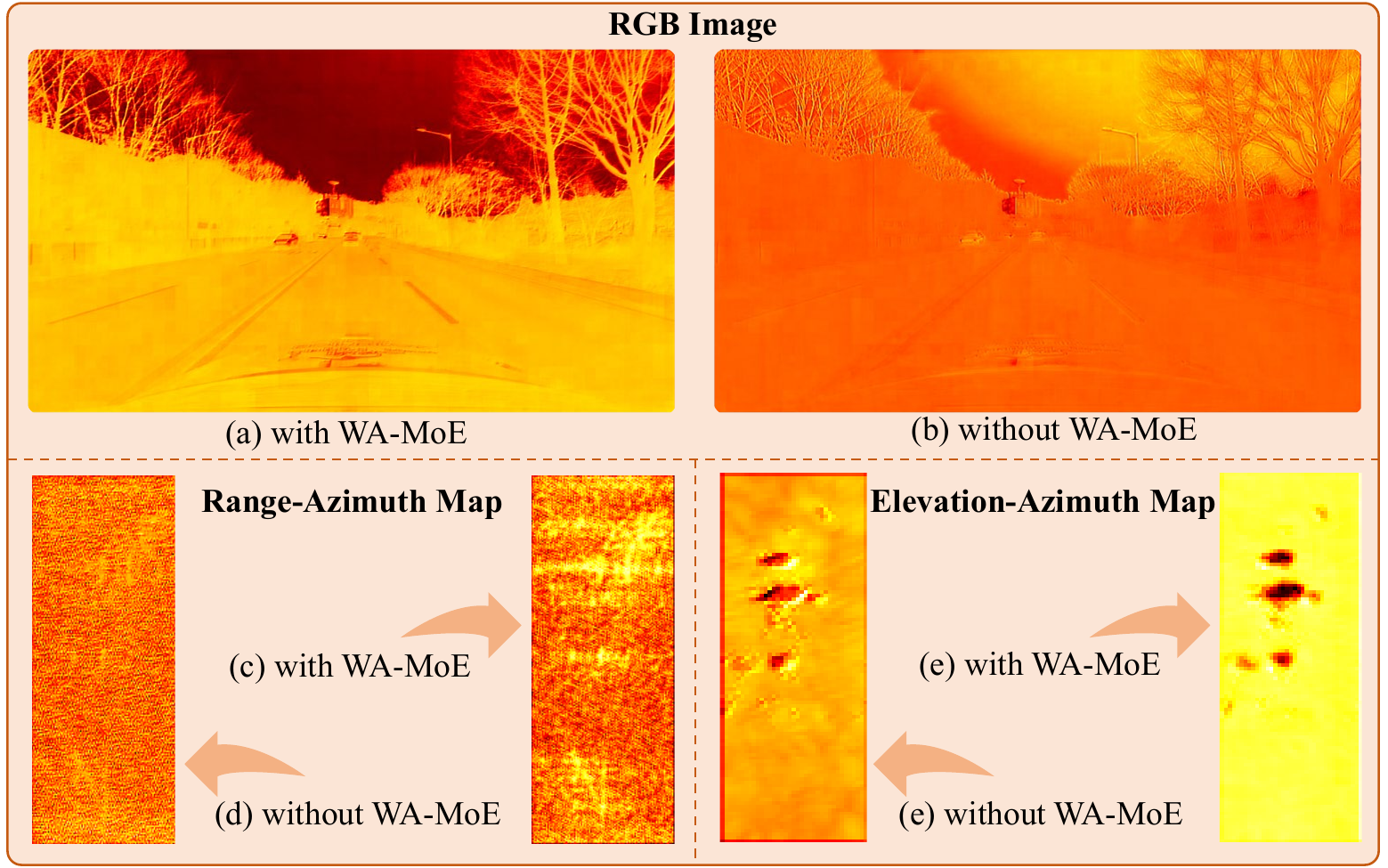}
    \vspace{-3mm}
    \caption{Visualization of feature maps with WA-MoE and without WA-MoE.}
    \label{fig:feature_vis}
\end{figure}

For the Wavelet Attention MoE (WA-MoE), we visualize in Fig. \ref{fig:feature_vis} the feature maps obtained with and without WA-MoE across three input modalities: RGB images, RA maps, and EA maps. The results show that the features enhanced by wavelet attention exhibit more pronounced object activations, whereas those from plain convolutional networks suffer from lower signal-to-noise ratios. This indicates that the wavelet‑attention mechanism effectively amplifies foreground responses while suppressing background noise.

\begin{figure}
    \centering
    \includegraphics[width=0.95\linewidth]{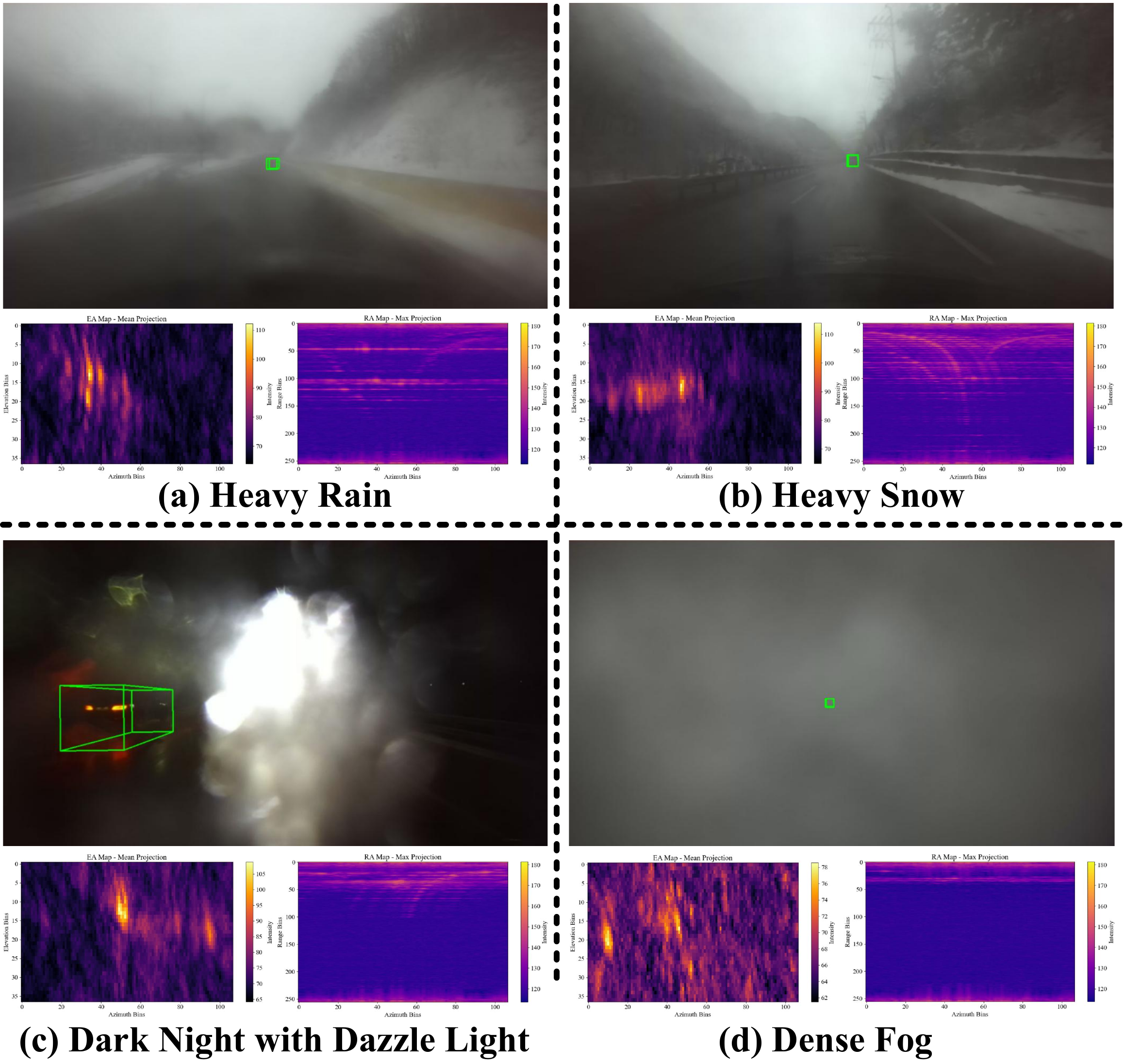}
    \vspace{-3mm}
    \caption{Visualization of prediction by WRCFormer under extremely adverse situations with the corresponding radar maps.}
    \label{fig:vis_adverse}
\end{figure}

Fig.\ref{fig:vis_adverse} presents the results of WRCFormer under highly adverse conditions, demonstrating that WRCFormer maintains robust performance even in such scenarios. Alongside the detections, we visualize the corresponding RA and EA map. It is observed that the mmWave radar effectively penetrates rain, fog, and snow, yielding clear target returns, whereas the targets are nearly invisible in the RGB images. This comparison strongly validates the critical advantage and indispensability of radar in perception under extreme weather conditions.

\section{Conclusion}
We introduced WRCFormer, a robust 3D object detector that effectively fused visual features with raw 4D radar tensors. The proposed method introduced two key innovations: a Wavelet Attention Mixture-of-Experts module that captures joint spatial-frequency representations to enhance sparse radar signatures, and a two-stage Geometry-guided Progressive Fusion mechanism that enables efficient and semantically-aligned cross-modal fusion integration. Extensive experiments on the K-Radar dataset proved that WRCFormer set a new state-of-the-art performance, with particularly strong robustness under adverse weather conditions where camera-based perception often degrades. These results underscored the significant advantage of directly leveraging rich 4D radar tensor, paired with specialized neural architectures, for building reliable, all-weather perception systems in autonomous driving.


%





\ifCLASSOPTIONcaptionsoff
  \newpage
\fi



%
\normalem
\footnotesize
\bibliographystyle{IEEEtran}
\bibliography{ref}

%
\newpage
\end{document}